\documentclass{article}

\PassOptionsToPackage{square,numbers}{natbib}
\usepackage[preprint]{neurips_2020}

\usepackage[utf8]{inputenc} 
\usepackage[T1]{fontenc}    
\usepackage[colorlinks]{hyperref}       
\usepackage{url}            
\usepackage{booktabs}       
\usepackage{amsfonts}       
\usepackage{nicefrac}       
\usepackage{microtype}      

\usepackage{amssymb}
\usepackage{bm}
\usepackage{adjustbox}
\usepackage{multirow}
\usepackage{mathtools}
\usepackage{subcaption}
\usepackage{lineno}
\usepackage{setspace}
\usepackage{enumitem}
\usepackage{amsmath}
\usepackage{comment}
\usepackage{bbm}
\usepackage[table]{xcolor}
\usepackage{authblk}
\usepackage{caption}

\usepackage{booktabs}

\newcommand{\R}{\mathbb{R}} 
\newcommand{\E}{\mathbb{E}} 
\newcommand{\N}{\mathcal{N}} 
\newcommand{\M}{\mathcal{M}} 
\DeclareMathOperator{\Tr}{Tr} 

\DeclarePairedDelimiter\p{(}{)}
\DeclarePairedDelimiter\cor{[}{]}
\DeclarePairedDelimiter\llav{\{}{\}}
\DeclarePairedDelimiter\ang{\langle}{\rangle}

\DeclareMathOperator{\Xm}{\mathbf{X}^{(m)}} 
\DeclareMathOperator{\XmT}{\mathbf{X}^{(m)^T}} 
\DeclareMathOperator{\Xnm}{\mathbf{x}_{n,:}^{(m)}} 
\DeclareMathOperator{\Xonem}{\mathbf{x}_{1,:}^{(m)}} 
\DeclareMathOperator{\XNm}{\mathbf{x}_{N,:}^{(m)}} 
\DeclareMathOperator{\Xndm}{x_{n,d}^{(m)}} 

\DeclareMathOperator{\Km}{\mathbf{K}^{(m)}}
\DeclareMathOperator{\KmT}{\mathbf{K}^{(m)^T}}
\DeclareMathOperator{\Knm}{\mathbf{k}_{n,:}^{(m)}}

\DeclareMathOperator{\Knnsm}{k_{n,n}^{(m)}}
\DeclareMathOperator{\Knum}{k_{n,u}^{(m)}}
\DeclareMathOperator{\KM}{\mathbf{K}^{\{\M\}}} 
\DeclareMathOperator{\KnM}{\mathbf{k_{n,:}}^{\{\M\}}} 

\DeclareMathOperator{\lm}{\text{\boldmath$\lambda$}^{(m)}} 
\DeclareMathOperator{\lonem}{\lambda_1^{(m)}} 
\DeclareMathOperator{\lDm}{\lambda_{D_m}^{(m)}} 
\DeclareMathOperator{\ldm}{\lambda_d^{(m)}} 

\DeclareMathOperator{\Z}{\mathbf{Z}} 
\DeclareMathOperator{\ZT}{\mathbf{Z}^T} 
\DeclareMathOperator{\Zn}{\mathbf{z}_{n,:}} 
\DeclareMathOperator{\ZnT}{\mathbf{z}_{n,:}^T} 

\DeclareMathOperator{\Am}{\mathbf{A}^{(m)}} 
\DeclareMathOperator{\AmT}{\mathbf{A}^{(m)^T}} 
\DeclareMathOperator{\Akm}{\mathbf{a}_{:,k}^{(m)}} 
\DeclareMathOperator{\Anm}{\mathbf{a}_{n,:}^{(m)}} 
\DeclareMathOperator{\Aum}{\mathbf{a}_{u,:}^{(m)}} 
\DeclareMathOperator{\AumT}{\mathbf{a}_{u,:}^{(m)^T}} 
\DeclareMathOperator{\Ankm}{a_{n,k}^{(m)}} 

\DeclareMathOperator{\Wm}{\mathbf{W}^{(m)}} 
\DeclareMathOperator{\WmT}{\mathbf{W}^{(m)^T}} 
\DeclareMathOperator{\Wkm}{\mathbf{w}_{:,k}^{(m)}} 

\DeclareMathOperator{\am}{\text{\boldmath$\alpha$}^{(m)}} 
\DeclareMathOperator{\akm}{\alpha_k^{(m)}} 

\DeclareMathOperator{\gamm}{\text{\boldmath$\gamma$}^{(m)}} 
\DeclareMathOperator{\gamnm}{\gamma_{n}^{(m)}} 

\DeclareMathOperator{\taum}{\tau^{(m)}} 

\DeclareMathOperator{\sumd}{\sum\limits_{d=1}^{D_m}} 
\DeclareMathOperator{\sumn}{\sum\limits_{n=1}^{N}} 
\DeclareMathOperator{\sumns}{\sum\limits_{\tilde{\tilde{n}}=1}^{\tilde{N}}} 
\DeclareMathOperator{\sumu}{\sum\limits_{u=1}^{N}} 
\DeclareMathOperator{\summ}{\sum\limits_{m=1}^{M}} 
\DeclareMathOperator{\sumk}{\sum\limits_{k=1}^{K_c}} 

\DeclareMathOperator{\prodn}{\prod\limits_{n=1}^{N}} 
\DeclareMathOperator{\prodk}{\prod\limits_{k=1}^{K_c}} 

\newcommand{\lnp}[1]{\ln\p*{#1}} 

\newcommand{\eqeq}{\enskip=&\enskip}
\newcommand{\eqapprox}{\enskip\approx&\enskip}
\newcommand{\eqsimil}{\enskip\sim &\enskip}

\usepackage{ulem}
\usepackage{xcolor}

\title{Bayesian Sparse Factor Analysis with Kernelized Observations}

\author{Carlos Sevilla-Salcedo\thanks{Corresponding author. Email address: sevisal@tsc.uc3m.es}, Alejandro Guerrero-L\'{o}pez, Pablo M. Olmos\thanks{Pablo M. Olmos is also with the Gregorio Mara\~n\'on Health Research Institute.}, Vanessa G\'{o}mez-Verdejo}
\affil{\small Department of Signal Processing and Communications, Universidad Carlos III de Madrid Legan\'es, 28911 Spain}

\begin{document}

\maketitle

\begin{abstract}
Multi-view problems can be faced with latent variable models since they are able to find low-dimensional projections that fairly capture the correlations among the multiple views that characterise each datum. On the other hand, high-dimensionality and non-linear issues are traditionally handled by kernel methods, inducing a (non)-linear function between the latent projection and the data itself. However, they usually come with scalability issues and exposition to overfitting. Here, we propose merging both approaches into single model so that we can exploit the best features of multi-view latent models and kernel methods and, moreover, overcome their limitations.

In particular, we combine probabilistic factor analysis with what we refer to as kernelized observations, in which the model focuses on reconstructing not the data itself, but its relationship with other data points measured by a kernel function. 
This model can combine several types of views (kernelized or not), and it can handle heterogeneous data and work in semi-supervised settings. Additionally, by including adequate priors, it can provide compact solutions for the kernelized observations -based in a automatic selection of Bayesian Relevance Vectors (RVs)- and can include feature selection capabilities. Using several public databases, we demonstrate the potential of our approach (and its extensions) w.r.t. common multi-view learning models such as kernel canonical correlation analysis or manifold relevance determination.

\end{abstract}
\setcounter{footnote}{0} 

\section{Introduction}
Given a set of observable data, Latent Variable Models (LVMs) \citep{loehlin1987latent} aim to extract a reduced set of hidden variables able to summarise the information into a low dimensional space. 
These models have become crucial in multi-view problems \citep{atrey2010multimodal, sharma2012generalized, li2019generative}, where data is represented by different modalities or views, since LVMs are able to explain the common information among all the modalities. 
One of the most well-known types of LVMs are the
classical MultiVariate Analysis (MVA) methods, such as Principal Component Analysis (PCA) and Canonical Correlation Analysis (CCA) \citep{pearson1901liii,hotelling1936relations}, which aim to exploit the data correlation to obtain a low dimensional latent representation of the data. Its usage has been generalised due to its easy non-linear extension by means of kernel methods \citep{scholkopf1998nonlinear, zhang2016multi}. The fact of supporting a kernel formulation allows these methods to learn arbitrarily complex non-linear models with a complexity determined by the number of training points \citep{yu2013kernel} and make them highly convenient in scenarios with high dimensional data.

Probabilistic formulations of LVMs methods  are known as Factor Analysis (FA) \citep{harman1976modern}, which emerge as a linear Bayesian framework where one can obtain the desired latent representation together with a measure of the uncertainty. Among their many variants, such as Probabilistic PCA \citep{tipping1999probabilistic}, Supervised PCA \citep{yu2006supervised}, Bayesian Factor Regression \citep{bernardo2003Bayesian} or Bayesian CCA \citep{klami2013bayesian}, the Inter-Battery FA model \citep{klami2013bayesian} stands out for its capability of handling not only latent variables associated to the common information among all the views, but also for being able to model the intra-view information. This model has been recently extended in \citep{sevilla2020sparse}, named as Sparse Semi-supervised Inter-Battery Bayesian Analysis (SSHIBA),  to incorporate missing attributes, feature selection and the ability to handle heterogeneous data such as categorical or multi-dimensional binary data.


The use of kernel methods in Bayesian approaches has been mostly developed with Gaussian Processes (GP) \citep{williams2006Gaussian} and their non-supervised version to perform dimensionality reduction explored in GPLVMs \citep{lawrence2005probabilistic}. These approaches combine the advantages of the kernels methods, exploiting the non-linear relationships among the data, with those of a probabilistic framework. In \citep{damianou2012manifold}, the authors propose a shared GPLVMs approach, called Manifold Relevance Determination (MRD), to provide a non-linear latent representation for multi-view learning problems. This model is extended in \citep{damianou2016multi}, including an Automatic Relevance Determination (ARD) prior \citep{neal2012Bayesian} over the kernel formulation, to endow it with feature relevance analysis.

GPLVMs come with practical scalability drawbacks that need to be addressed. The cubic complexity with the number of training points requires the use of inducing points and variational approaches \citep{pmlr-v5-titsias09a}. Selecting the number of inducing points to use, and where to place them in the latent space, is still a challenging problem, being a common solution to place them in a regular basis along the latent space and only optimize the pseudo-observation at those points \citep{pmlr-v37-wilson15}. Furthermore, up to our knowledge, there is no versatile implementation in the state-of-the-art of a multi-view GPLVM able to handle heterogeneous observations (integer, categorical, real and positive observations) and missing values.

In this paper we propose a novel method to implement non-linear probabilistic LVMs that still builds upon a linear generative model, hence inheriting their computational and scalability properties. Instead of implementing a kernel method, i.e. a GP, to move from the latent representation to the observed data, we propose to reformulate probabilistic FA so that it generates kernel relationships instead of data observations. In the same way that Kernelized PCA (KPCA) or Kernelized CCA (KCCA) are able to generate non-linear latent variables by linearly combining element of a kernel vector; here, from a Bayesian generative point of view, we first i.i.d. sample latent representations and 
project on an $N$-dimensional space (being $N$ the number of points) using a weight matrix representing the dual parameters.
We apply this trick over the SSHIBA formulation \citep{sevilla2020sparse}
to exploit their functionalities over this kernelized formulation. Thanks to that, we can efficiently face semi-supervised heterogeneous multi-view problems combining linear and non-linear data representations; in this way, 
one can combine kernelized views to deal with non-linear relationships with linearly kernelized to work with high dimensional problems.
Although \citep{lian2015integrating} presents a first attempt of using kernelized data representations for FA models, here we propose a complete framework including the  following novelties/functionalities: (1) we can work with any type of kernel and are not limited to linear kernels; (2) we can force the automatic selection of Relevance Vectors (RVs) of the kernel to obtain scalable solutions  (equivalent to the support vectors of the Support Vector Machines (SVMs)); (3) we can include an ARD kernel to obtain a per-variable relevance that can be exploited for feature selection; (4) we can adapt the multiview model to work as multiple kernel learning algorithm for FA models.


The article is organised as follows: Section \ref{sec:KSSHIBA} presents the kernelized formulation of SSHIBA as well as the proposed formulations for the RVs and feature selection. Section \ref{sec:Results} defines the implementation details and the baselines and databases used for the experiments in different scenarios, where we show the performance of each proposed extension to the model. Finally, Section \ref{sec:conclusion} gives some final conclusions and highlights the main results. 

\section{Bayesian Sparse Factor Analysis with Kernelized Observations}
\label{sec:KSSHIBA}

Let's consider a multi-view problem where we have $N$ data samples represented in $M$ different modalities, $\llav{\Xm}_{m=1}^M$, and our goal is to find an inter and intra-view non-linear latent representation, $\Z$. That is, given that $\Xnm \in \R^{D_m}$ is the $m$-th view   of  the $n$-th data, $\Zn$ has to compress, in a low dimensional space of size $K_c <<( D_1, \ldots, D_M)$ both the intra-view and inter-view information
of $\Xnm$ over all the views exploiting the correlations among the data.\footnote{Given a matrix $\mathbf{B}$, we denote the $i$-th row by $\mathbf{b}_{i,:}$ and the $j$-th column by $\mathbf{b}_{:,j}$.}
Whereas kernel MVA obtains the latent representation of the $n$-th data as a linear combination, by some dual variables, of its kernel representation $\Knm$, here we propose to reformulate this idea from a generative point of view.
In particular, we start from the SSHIBA algorithm formulation \citep{sevilla2020sparse} (see \ref{sec:AppendixSSHIBA} for a brief introduction) and consider that there exist some latent variables $\Zn \sim \N(\bf{0},\bf{I}_{K_c})$ which are linearly combined with a set of dual variables $\Am \in \R^{N \times K_c}$ to generate a kernel vector, $\Knm$, as:
\begin{align}
\Knm \eqeq \Zn \AmT + \taum \label{eq:KernelDefinition}
\end{align}
where $\taum$ is zero-mean Gaussian noise, with noise power following a Gamma distribution of parameters $a^{\taum}$ and $b^{\taum}$, 
and, given a mapping function $\phi(\cdot)$ and its associated kernel function $K(\bf x, \bf x') = \phi(\bf x)^\top  \phi(\bf x') $, $\Knm$ is the kernel between $\Xnm$ and all the training data, i.e.,  $\Knm = \cor{K\p{\Xnm,\Xonem}, \dots, \\ K\p{\Xnm,\XNm}}$. The dual variable matrix $\Am$ plays the role of the linear projection matrix and it is defined using the same structured ARD prior considered in both \citep{klami2013bayesian} and \citep{sevilla2020sparse}. Namely, an ARD prior that promotes that full rows of this matrix are cancelled, i.e.
\begin{align}
    \Akm \eqsimil \N\p*{0,\p*{\akm}^{-1} I_{K_c}} \label{eq:Adist}\\
    \alpha_{k}^{(m)} \eqsimil \Gamma\p*{a^{\alpha^{(m)}}, b^{\alpha^{(m)}}},
\end{align}
so that the product in \eqref{eq:KernelDefinition} induces sparsity in the latent factors, leading to a selection of the appropriate set of them \citep{tipping2001sparse}.

\begin{figure}[t]
 \centering
 \includegraphics[page=1,width=\linewidth]{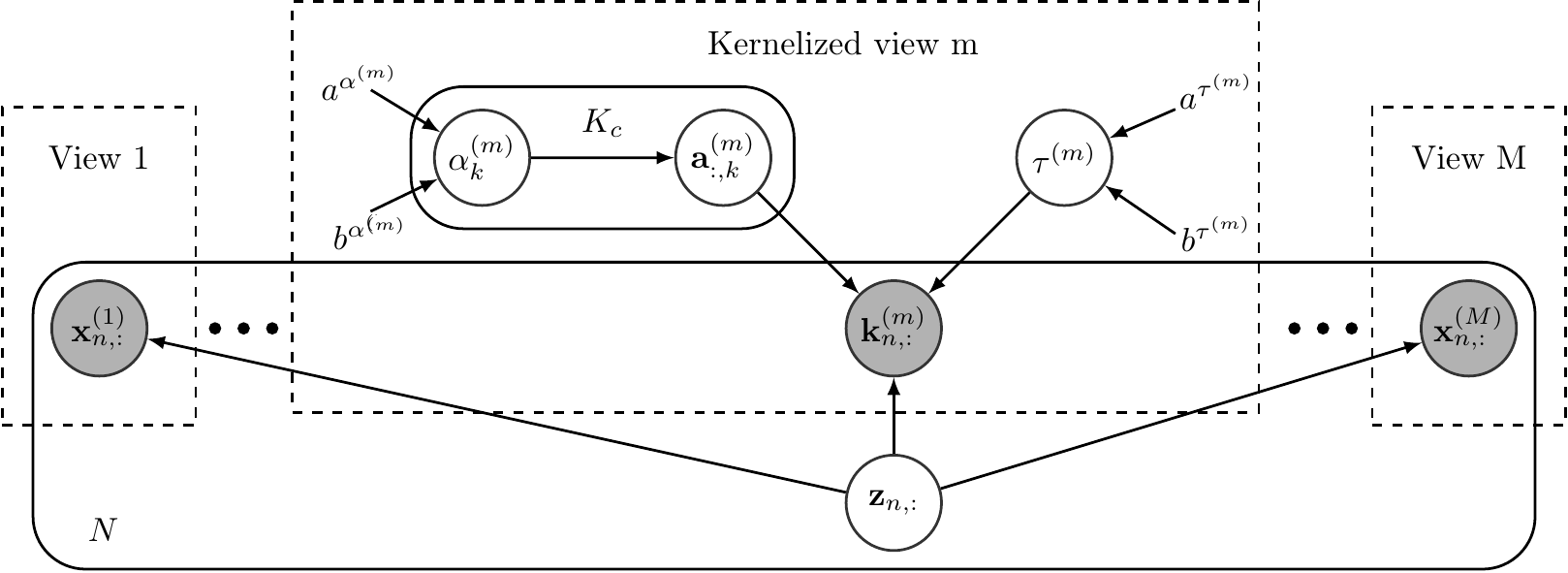}\\
 \captionof{figure}{Graphic model of SSHIBA with kernelized views (KSSHIBA).}
  \label{fig:Scheme_kernel}
\end{figure}

Figure \ref{fig:Scheme_kernel} shows the graphical model of KSSHIBA. 
Following \citep{klami2013bayesian}, for the data views that are directly explained given the latent projection 
we have $\Xnm = \Zn\WmT+\taum$, where the weight matrix $\Wm$ follows the same structured ARD prior mentioned above. We can refer to these as primal observations. For some other views, we might be interested in explaining them indirectly through a kernelized observation following \eqref{eq:KernelDefinition}. This conversion can be of interest when either we want to exploit intra and inter-view non-linear relationships, or when the view's dimensionality is much larger than the number of data points $N$ and we prefer to work in the dual space, to reduce the number of parameters to be learnt. When both primal and kernelized observations are used, the learned latent projection $\Zn$  attempts to faithfully reconstruct the original data representation in the primal views, as well as their kernel representation, i.e., their (non)-linear relationships with the other data points.

\begin{figure}[t]
  \centering
  \rotatebox{90}{\hspace{0.5cm} \small Linear Kernel}
  \begin{subfigure}[t]{0.28\linewidth}
    \centering
    \includegraphics[width=\linewidth]{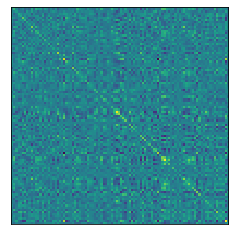}\\
  \end{subfigure}
  ~
  \begin{subfigure}[t]{0.28\linewidth}
    \centering
    \includegraphics[width=\linewidth]{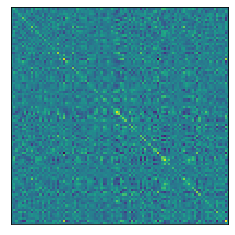}
  \end{subfigure}
  ~
    \begin{subfigure}[t]{0.34\linewidth}
    \centering
    \includegraphics[width=\linewidth]{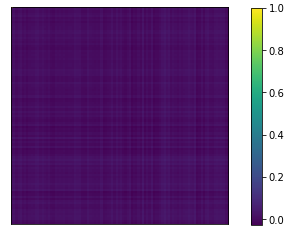}
  \end{subfigure}
  \\
  \rotatebox{90}{\hspace{0.8cm} \small RBF Kernel}
  \begin{subfigure}[t]{0.28\linewidth}
    \centering
    \includegraphics[width=\linewidth]{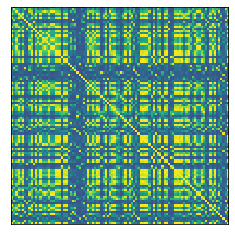}\\
    \caption{Original kernel matrix}
    \label{fig:original_rbf}
  \end{subfigure}
  ~
  \begin{subfigure}[t]{0.28\linewidth}
    \centering
    \includegraphics[width=\linewidth]{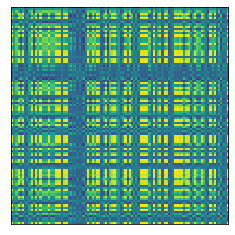}
    \caption{Reconstructed matrix}
    \label{fig:learnt_rbf}
  \end{subfigure}
    ~
  \begin{subfigure}[t]{0.34\linewidth}
    \centering
    \includegraphics[width=\linewidth]{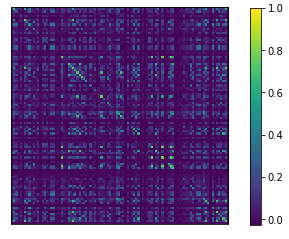}
    \caption{Reconstruction error}
    \label{fig:difference_rbf}
  \end{subfigure}
    \caption{Example of the generative properties of KSSHIBA to reconstruct a complete kernel matrix.}
  \label{fig:kernel}
\end{figure}

Note that sampling from the model in \eqref{eq:KernelDefinition} does not ensure a valid kernel positive semi-definite matrix. The kernel matrix is simply treated as an observation (a kernelized observation) and, as such, the model parameters will be chosen to minimize the reconstruction error. In Figure \ref{fig:kernel} we include a graphical representation of both a kernelized observation and the map reconstruction through \eqref{eq:KernelDefinition} using the  mean of the posterior distribution of $\Zn$, where we can observe that the kernelized observations (or kernel matrix) are almost perfectly reconstructed. Certainly, more appropriate models could be used to adapt the observation model (given $\Zn$) to the properties of a kernelized observation. To address this issue, we have explored alternative formulations based in non-independent noise; for example, defining the noise distribution as an inverse-wishart to have a full rank covariance noise or modelling its covariance as the product of two low rank matrices \citep{murphy2012machine}. However, these schemes led to considerably more complicated (less flexible) formulation which limited the rest of the properties of this proposal (as the ones proposed in the following sections). Henceforth, in this work, we restrict to the model in \eqref{eq:KernelDefinition}, and leave this line of work open for future research.

%
%

With this model we can evaluate the posterior distribution of the variables given the observed data
\begin{align}\label{eq:post}
p(\Theta|\mathbf{k}_{1,:},\dots, \mathbf{k}_{N,:}) &=\frac{\prod_{n=1}^{N}p(\mathbf{k}_{n,:}|\Theta)p(\Theta)}{p\p*{\mathbf{k}_{1,:},\dots, \mathbf{k}_{N,:}}},\\\label{eq:marglik}
p\p*{\mathbf{k}_{1,:},\dots, \mathbf{k}_{N,:}} &= \int p\p*{\Theta, \mathbf{k}_{1,:},\dots, \mathbf{k}_{N,:}} d(\Theta),
\end{align}
where $\Theta$ comprises all random variables (rv) in the model. As presented in \citep{sevilla2020sparse}, we rely on an approximate inference approach through mean-field variational inference \citep{Blei17}, where a lower bound to \eqref{eq:marglik} of the form
\begin{align}\label{elbo}
\log p\p*{\mathbf{k}_{1,:},\dots, \mathbf{k}_{N,:}} \geq \int q(\Theta) \log\left(\frac{\prod_{n=1}^{N}p(\mathbf{k}_{n,:}|\Theta)p(\Theta)}{q(\Theta)}\right)d(\Theta)
\end{align}
is maximised, and a fully factorised variational family is chosen to approximate the posterior distribution in \eqref{eq:post}
\begin{align}
p(\Theta|\KM)
\eqapprox \prod_{m=1}^{\M}\p*{q\p*{\Am} q\p*{\taum} \prodk q\p*{\akm}} \prod_{n=1}^{N}q\p*{\Zn}  \label{eq:qModel}
\end{align}
where $\KM$ represents a stacked version of the kernel for each sample, $\KnM$, of dimension $N\times D_m$.

The mean-field posterior structure along with the lowerbound in \eqref{elbo} results into a feasible coordinate-ascent-like optimization algoritm in which the optimal maximization of  each of the factors in \eqref{eq:qModel} can be computed if the rest remain fixed using the following expression
\begin{align}\label{eq:meanfield}
q^*(\theta_i) \propto \mathbb{E}_{\Theta_{-i}}\left[\log p(\Theta,\mathbf{k}_{1,:},\dots, \mathbf{k}_{N,:})\right],
\end{align}
where $\Theta_{-i}$ comprises all rv but $\theta_i$. This new formulation is in general feasible since it does not require to completely marginalize $\Theta$ from the joint distribution. 


Redefining the input matrix as a kernel matrix leads to a formulation equivalent to SSHIBA, which, accordingly, permits to keep the previous model functionalities. Table \ref{tab:KSSHIBA} shows the KSSHIBA mean-field factor update rules. For a compact notation, we stuck in matrix $\mathbf{Z}$, of dimension $N\times K_c$, the latent projection of all data points and $<>$  represents the mean value of the rv. 
The posterior distribution of all model parameters and latent projections is approximated using variational inference with a fully factorized posterior where it can be noted that each update has a computational cost of $O(N^2K_c+K_c^3)$.

\begin{table}[h!b]
\caption{Updated rules, obtained by a mean field approximation, of $q$ distribution for the different variables of KSSHIBA model.}
\begin{adjustbox}{max width=\textwidth}
\renewcommand{\arraystretch}{1.}
\centering
\begin{tabular}{ccc}
\toprule
{\textbf{Variable}} & { $\bm{q}^*$ \textbf{distribution}} & \textbf{Parameters} \\\midrule
\multirow{3}{*}{$\Zn$}
& \multirow{3}{*}{$\N\Zn | \mu_{\Zn},\Sigma_{\Z}$}
& $\mu_{\Zn} = \summ\p*{\ang{\taum} \Km \ang{\Am} \Sigma_{\Z}}$ \\
& & $\Sigma_{\Z}^{-1} = \p*{I + \summ\p*{\ang{\taum} \ang{\AmT \Am}}}$ \\&&\\
\midrule                     
\multirow{3}{*}{$\Am$} 
& \multirow{2}{*}{$\prodn \p*{\N \p*{\Anm | \mu_{\Anm}, \Sigma_{\Am}}}$} 
& $\mu_{\Anm} = \ang{\taum} \KmT \ang{\Z} \Sigma_{\Am}$ \\
& & $\Sigma_{\Am}^{-1} = \p*{\text{diag}(\ang{\am}) + \ang{\taum}\ang{\ZT \Z}}$  \\&&\\
\midrule    
\multirow{3}{*}{$\akm$}
& \multirow{2}{*}{$\Gamma\p*{\akm | a_{\akm},b_{\akm}}$}
& $a_{\akm} = \frac{D_m}{2} + a^{\am}$ \\ 
& & $b_{\akm} = b^{\am} + \frac{1}{2} \ang{\AmT\Am}_{k,k}$ \\&&\\
\midrule
\multirow{5}{*}{{$\taum$}}
& \multirow{5}{*}{$\Gamma\p*{\taum | a_{\taum},b_{\taum}}$}
& $a_{\taum} = \frac{D_m N}{2} + a^{\taum}$ \\
& & $b_{\taum} = b^{\taum} + \frac{1}{2} \left(\sumn\sumns \Knnsm^2 \right.$ \\
& & $\left. - 2 \Tr\llav*{\ang{\Am}\ang{\ZT}\Km}\right.$  \\
& & $\left. + \Tr\llav*{\ang{\AmT\Am} \ang{\ZT \Z}}\right)$  \\
\bottomrule
\end{tabular}
\end{adjustbox}
\label{tab:KSSHIBA}
\end{table}

\subsection{Multiple kernel learning}


By using different kernels in different views, the model adapts to a multiple kernel learning scenario (MKL SSHIBA). To further analyse the implications of this combination, let's analyse the update rule for the mean of latent variables $\Z$
\begin{align}\label{eq:meanZ}
\mu_{\Zn} = \summ \ang{\taum} \Km \ang{\Am} \Sigma_{\Z}.
\end{align}
Here we can see that the latent factors are obtained as a linear combination of the kernels on each view. This entails the additional advantage of automatically learning these mixing coefficients of this combination (depending on parameters $\taum$, $\Am$ and $\Sigma_{\Z}$) through variational inference. Furthermore, if we define one view as an output view, the values that need to be predicted are calculated using this learnt variable Z, therefore conserving the MKL structure for the prediction.

Most MKL methods are based on heuristics \citep{de2010methods, qiu2008framework} or two step optimization process \citep{fung2004fast}. However, the proposed model determines the kernel parameters trhough the mean-field update. Furthermore, this formulation also benefits from the advantages of the SSHIBA formulation, namely semi-supervised learning, feature selection and RV selection.





\subsection{Automatic Bayesian Relevance Vector Selection}

On the basis of a full $N\times N$ kernel, with a more structured ARD prior we can achieve not only the shrinkage of the number of effective latent factors, but also a more compact representation of the data by means of a reduced kernel matrix in which only a reduced set of RVs are kept. 

For this purpose, the proposed formulation can introduce a double ARD prior over the dual variables $\Am$, 
\begin{align}
    \Ankm \eqsimil \N\p*{0,\p*{\gamnm\akm}^{-1}} \label{eq:Aspdist}
\end{align}
This way, $\akm$ continues forcing row-wise sparsity to automatically select the number of latent factors and, additionally, 
\begin{align}
    \gamnm \eqsimil \Gamma\p*{a^{\gamm}, b^{\gamm}} \label{eq:Gamdist}
\end{align}
induces column-wise sparsity, which automatically forces many of the RVs to not influence in the final model, consequently achieving a more compact solution.
Table \ref{tab:KSSHIBA_sparse} shows the KSSHIBA mean-field factor update rules for the double ARD case.

\begin{table}[h!b]
\caption{Updated $q$ distribution for the automatic RV selection.}
\begin{adjustbox}{max width=\textwidth}
\renewcommand{\arraystretch}{1.}
\centering
\begin{tabular}{ccc}
\toprule
\textbf{Variable} & {$\bm{q}^*$ \textbf{distribution}} & \textbf{Parameters} \\\midrule
{\multirow{2}{*}{$\Am$}}
& \multirow{2}{*}{$\prodn\N \p*{\Anm | \mu_{\Anm}, \Sigma_{\Anm}}$}
& $\mu_{\Am} = \ang{\taum} \XmT \ang{\Z} \Sigma_{\Wm}$ \\
& & $\Sigma_{\Anm}^{-1} = \text{diag}(\ang{\am})\ang{\gamnm} + \ang{\taum}\ang{\ZT \Z}$ \\
\midrule
\multirow{3}{*}{$\gamm$}
& \multirow{3}{*}{$\prodn \Gamma\p*{\gamnm | a_{\gamnm},b_{\gamnm}}$}
& $a_{\gamnm} = \frac{K_c}{2} + a^{\gamm}$ \\
& & $b_{\gamnm} = b^{\gamm} + \frac{1}{2} \sumk\ang{\akm}\ang{\Ankm\Ankm}$ \\
\bottomrule
\end{tabular}
\end{adjustbox}
\label{tab:KSSHIBA_sparse}
\end{table}
%
%
%
%
%

\subsection{Automatic feature-relevance determination}\label{sec:fl}

Furthermore, we can additionally endow the proposed kernelized data representation with automatic feature relevance. If by using the double ARD structure we can cancel full rows or columns, equivalently, by using an ARD kernel we can perform this feature relevance determination. In the ARD kernel, each feature of the original observations is multiplied by a variable $\ldm$ in the kernel definition. For example, for a Radial Basis Function (RBF) kernel,
\begin{align}
    \Knnsm \eqeq  \exp\p*{-\sumd\p*{\Xndm - \Xndm}^2 \ldm}, \label{eq:sparseKrbf}
\end{align}
we can optimise $\lm = \cor{\lonem, \dots, \lDm}$ by maximising the lower bound of our mean field approach given by direct optimisation over the variational lower bound. In our model, if the $m$-th view is kernelized then the only term in the lower bound affected by the ARD is the $\E_{q}\cor*{\lnp{p\p*{\Km|\Theta}}}$, having that (see \citep{sevilla2020sparse} for further details on the lower bound):
\begin{align}
    LB \eqeq  -\frac{\ang{\taum}}{2} \sumn\sumu\p*{\Knum^2 - 2\Knum \ang{\Aum} \ang{\ZnT} + \ang{ \AumT,\Aum} \ang{ \ZnT,\Zn}} \label{eq:LBderivative}
\end{align}
We alternate between mean-field updates over the variational bound and direct maximization of \eqref{eq:LBderivative} w.r.t. $\lm$ using any gradient ascend method (we use Pytorch and Adam for such updates). Finally, by setting a threshold for $\lm$, the model is capable of automatically selecting the most relevant features while training.

\section{Results}
\label{sec:Results}

Throughout this section we define the different baselines as well as the implementation details for the analysis of the methods in terms of performance and interpretability of the inferred model parameters and latent projections. 
Furthermore, an exemplary notebook with the \textit{Python} code of the proposed method is openly available in this github repository: \href{https://github.com/sevisal/KSSHIBA.git}{KSSHIBA}.

\subsection{Experimental setup}
\label{sec:Baselines}
In order to truthfully analyse the performance of the proposed framework, we decided to compare it with some state-of-the-art algorithms with similar capabilities. In particular we wanted to focus on some of the most relevant algorithms for FA. For the multi-dimensional regression datasets, we compare ourselves with KPCA, KCCA, MRD, Support Vector Regression (SVR) and MultiLayer Perceptron (MLP). Conversely, for the classification datasets, we compare with CCA and Support Vector Machine (SVM). We will now explain the specific parameters and formulations we used for these baselines.

As we proposed a kernel version of a FA model, we decided to compare to most common non-linear LVM methods, such as, KPCA and KCCA with a RBF kernel. For each of these models we explored 20 values of $\gamma$ kernel hyperparameter in log-scale from $[10^{-8}$ to $10^{0.5}]$ divided by the number of tasks ($C$). In order to carry out predictions KPCA was combined with Linear Regression (KPCA+LR), as KCCA can work in a supervised manner, we decided to use it on its own and with LR (KCCA+LR) for a complete comparative analysis. The number of latent factors has been fixed to the maximum possible, $C$, in KCCA and to those which explain $95\%$ of the variance in KPCA. 

Due to the equivalency to some of our functionalities, namely working with multiview data, carrying out RV selection and latent representations,
we compared our model with the MRD \citep{damianou2012manifold} including an ARD for RVs selection. We used the available library in \textit{Matlab} \citep{damianou2012manifold}, setting the 
number of latents to twice the number of tasks ($2*C$). We used the RBF kernel with ARD and set the number of model optimisation iterations to 100 due to the long computational time required to train. 

Another kernel model used as baseline is SVR, where we also used a RBF kernel, exploring the same values of $\gamma$ as on KCCA/KPCA. The regularization parameter, $\lambda$, has also been validated, exploring 11 values in a logarithmic scale from $[10^{-4}$ to $10^{4}]$.

Finally, we also wanted to compare our model with a MLP neural network in two different scenarios: (1) a MLP to force a bottleneck of dimension $C$ in the hidden layers to have the same dimensionality reduction as KCCA. For this scenario, we validated two different configurations: (i) one hidden layer with $C$ neurons and (ii) two hidden layer with $C$ neurons and number of features ($D_m$) neurons, respectively. (2) A second scenario without bottleneck is presented, validating between three configurations: (i) one hidden layer with 100 neurons, (ii) two hidden layers with 100 and 50 neurons, and (iii) three hidden layers with 100, 50 and 100 neurons.



Regarding KSSHIBA, we have used its semi-supervised capability to predict the output, using the test samples (without their targets) during the training and, later, predicting their labels with the mean of the posterior distribution. To determine the number of iterations of the inference process of KSSHIBA, we used a convergence criteria based on the evolution of the lower bound. In particular, we stop the algorithm either when $mean(LB[-101:-2]) > LB[-1](1 - 10^{-4})$, where $LB[-1]$ is the lower bound at the last iteration and $mean(LB[-101:-2]$ the mean value of the previous values of the lower bound, or when it reaches $10^{4}$ iterations. The KSSHIBA  models were randomly initialized 10 times, keeping the one with the best lower bound. KSSHIBA automatically prunes the latent factors using the ARD prior included in the projection matrix $\Wm$. 

We calculated the hyperparameters of each model with a nested 10-folds Cross-Validation (CV). The outer CV is used to divide the dataset into training and test partitions, while the inner CV is in charge of validation and, therefore, it divides the training partition into a second training set and a validation set. This way we were able to estimate the performance of the whole framework and, additionally, validate the model parameters.
We used the coefficient of determination (R2) to compare the performance of the different variations of the methods and to adjust the method hyperparameters, which were CV.


\subsection{Performance evaluation of KSSHIBA for multi-dimensional regression}\label{sec:mulan_performance}

This section aims to analyse the performance of KSSHIBA for semi-supervised multi-dimensional regression in comparison with some state-of-the-art baselines. To do so, we used 9 multi-dimensional regression datasets from the \textit{Mulan} repository \citep{spyromitros2016multi, karalivc1997first, dvzeroski2000predicting}, whose main properties are summarized in Table \ref{MTR-datasets}.
\begin{table}[h!b]
\caption{Characteristic of the multi-task databases used in this work.}
\label{MTR-datasets}
\centering
\begin{adjustbox}{max width=\textwidth}
\begin{tabular}{cccc}
\toprule
Database & Samples & Features & Tasks \\ \midrule
\textit{at1pd}        & 337        & 411        & 6          \\
\textit{at7pd}        & 296        & 411        & 6          \\
\textit{oes97}        & 334        & 263        & 16         \\
\textit{oes10}        & 403        & 298        & 16         \\
\textit{edm}          & 154        & 16         & 2          \\
\textit{jura}         & 359        & 15         & 3          \\
\textit{wq}           & 1,060      & 16         & 14         \\
\textit{enb}          & 768        & 8          & 2          \\
\textit{slump}        & 103        & 7          & 3          \\ 
\bottomrule
\end{tabular}
\end{adjustbox}
\end{table}

Table \ref{tab:MultiTask} shows the results obtained by the proposed model in two different scenarios (one in which the number of latent factors $K_c$ is automatically learnt with the ARD prior and another in which we set $K_c$ to $ C$) and we compare their results with those of the baselines, namely, KCCA, KPCA, MRD, SVR and MLP.

\begin{table*}[h!b]
\caption{Results on multitask databases of KSSHIBA and the baselines. The white subrow represents the mean and standard deviation of R2 score and the gray subrow the number of effective latent factors found.}
\label{tab:MultiTask}
\centering
\begin{adjustbox}{max width=\linewidth}
\begin{tabular}{lcccccccccc}
\toprule
 & \multirow{2}{*}{KSSHIBA} & KSSHIBA & \multirow{2}{*}{MRD} & \multirow{2}{*}{KPCA + LR} & \multirow{2}{*}{KCCA + LR} & \multirow{2}{*}{SVR-RBF} & \multirow{2}{*}{MLP} \\ 
 &  &\small $K_c = C$ &  & &  &  &  \\\midrule
\multirow{2}{*}{\textit{at1pd}} 
& $\mathbf{0.79 \pm 0.09}$ & $0.78 \pm 0.09$ & $0.67 \pm 0.07$ & $0.67 \pm 0.12$ & $0.75 \pm 0.11$ & $0.01 $ & $0.77$\\
& \cellcolor{gray!10} $46 \pm 6$ & \cellcolor{gray!10} $6 $ & \cellcolor{gray!10} $12 $ & \cellcolor{gray!10} $22 \pm 10$ & \cellcolor{gray!10} $6 $ & $\pm 0.05$ & $\pm 0.11$\\ \midrule

\multirow{2}{*}{\textit{at7pd}} & $0.50 \pm 0.18$ & $0.52 \pm 0.13$ & $0.48 \pm 0.12$ & $0.39 \pm 0.19$ &  $\mathbf{0.57 \pm 0.16}$ & $0.01 $ & $0.35 $ \\
& \cellcolor{gray!10} $14 \pm 6$ & \cellcolor{gray!10} $6 $ & \cellcolor{gray!10} $12 $ & \cellcolor{gray!10} $21 \pm 1$ & \cellcolor{gray!10} $6 $ & $\pm 0.03$ & $\pm 0.69$  \\ \midrule

 \multirow{2}{*}{\textit{oes97}} & $\mathbf{0.71 \pm 0.10}$ & $0.69 \pm 0.10$ & $0.34 \pm 0.07$ & $0.45 \pm 0.20$ & $0.36 \pm 0.09$ & $0.39 $ & $0.58$ \\
& \cellcolor{gray!10} $17 \pm 6$ & \cellcolor{gray!10} $16 $ & \cellcolor{gray!10} $32 $ & \cellcolor{gray!10} $12 \pm 7 $ & \cellcolor{gray!10} $16 $ & $\pm 0.10$ & $\pm 0.21$\\ \midrule

 \multirow{2}{*}{\textit{oes10}} & $\mathbf{0.82 \pm 0.05}$ & $0.80 \pm 0.07$ & $0.38 \pm 0.07$  & $0.59 \pm 0.15$ & $0.43 \pm 0.12$ & $0.48 $ & $0.76 $ \\
 & \cellcolor{gray!10} $16 \pm 5$ & \cellcolor{gray!10} $16 $ & \cellcolor{gray!10} $32 $ & \cellcolor{gray!10} $14 \pm 7$ & \cellcolor{gray!10} $16 $ & $\pm 0.12$ & $\pm 0.08$\\ \midrule

 \multirow{2}{*}{\textit{edm}} & $\mathbf{0.51 \pm 0.18}$ & $0.21 \pm 0.09$ & $-0.17 \pm 0.45$ & $0.38 \pm 0.19$ & $0.18 \pm 0.26$ & $0.35 $ &  $0.26 $ \\
 & \cellcolor{gray!10} $30 \pm 11$ & \cellcolor{gray!10} $2 $ & \cellcolor{gray!10} $ $ & \cellcolor{gray!10} $16 \pm 5$ & \cellcolor{gray!10} $2 $ & $\pm 0.19$ &  $\pm 0.21$\\ \midrule

 \multirow{2}{*}{\textit{jura}} & $\mathbf{0.62 \pm 0.08}$ & $0.30 \pm 0.10$ & $0.57 \pm 0.06$ & $0.38 \pm 0.11$ & $0.18 \pm 0.15$ & $0.60 $ & $0.61$ \\
 & \cellcolor{gray!10} $21 \pm 3$ & \cellcolor{gray!10} $3 $ & \cellcolor{gray!10} $6 $ & \cellcolor{gray!10} $23 \pm 1$ & \cellcolor{gray!10} $3 $ & $\pm 0.05$ & $\pm 0.06$\\ \midrule

 \multirow{2}{*}{\textit{wq}} & $\mathbf{0.14 \pm 0.01}$ & $0.12 \pm 0.01$ & $-0.35 \pm 0.08$ & $0.09 \pm 0.02$ & $-0.01 \pm 0.01$ & $0.08 $ & $0.13$ \\
 & \cellcolor{gray!10} $76 \pm 9$ & \cellcolor{gray!10} $14 $ & \cellcolor{gray!10} $ 28 $ & \cellcolor{gray!10} $29 \pm 1$ & \cellcolor{gray!10} $14 $ & $\pm 0.02$ & $\pm 0.03$\\ \midrule

 \multirow{2}{*}{\textit{enb}} & $\mathbf{0.99 \pm 0.01}$ & $0.86 \pm 0.02$ & $0.91 \pm 0.01$ & $0.86 \pm 0.01$ & $0.98 \pm 0.01$ & $\mathbf{0.99}$ & $\mathbf{0.99 }$ \\
 & \cellcolor{gray!10} $118 \pm 4$ & \cellcolor{gray!10} $2 $ & \cellcolor{gray!10} $4 $ & \cellcolor{gray!10} $13 \pm 1$ & \cellcolor{gray!10} $2 $ & $\mathbf{\pm 0.01}$ & $\mathbf{\pm 0.08}$\\
\bottomrule
\end{tabular}
\end{adjustbox}
\end{table*}

In particular, we can see that KSSHIBA consistently outperforms most reference methods in every database, pointing out the performance advantages obtained in \textit{edm} and \textit{oes97}. Additionally, it is remarkable that this performance improvement is accomplished with an effective dimensional reduction, since KSSHIBA, applying a feature extraction, is able to outperform both a SVR and a MLP that use all the original features. At the same time, the results obtained by KSSHIBA with $K_c = C$ reveal that standard KSSHIBA is being too conservative in the number of extracted features and we could force a more restrictive pruning without degrading the final performance (note that  KSSHIBA with $K_c = C$ only deteriorates in the problems with only $2$ or $3$ (\textit{edm, jura} and \textit{enb}) since in this cases the number of latents is extremely reduced.


\subsection{Evaluation of the solution in terms of RVs}\label{sec:RV_exp}

Now, we want to test the capabilities of the KSSHIBA approach to automatically construct compact solutions by selecting a subset of training points, in other words, using RVs. For this purpose, we use the same databases and experimental setup as Section \ref{sec:mulan_performance}, but we compare with KPCA+LR and KCCA+LR using a Nyström \citep{nystrom} subsampling technique. In this way, KPCA+LR and KCCA+LR models will use this subsampling to select RV subsets; in these cases, the optimum percentage of RVs (respect to the total number of training data) has been selected by CV exploring their values in the set $[1,2,3,4,5,10,\dots,100] \%$.

Table \ref{tab:SupportVectors} shows 
 that the inclusion of the automatic RV selection on KSSHIBA keeps the original model performance for most databases, even improving it for \textit{oes97} and \textit{edm}. This is done while the model complexity is drastically reduced; in fact, analysing these results in detail, it is observed that the fact of reducing the number of RVs favours an additional reduction in the final number of latent factors. When comparing to KPCA+LR and KCCA+LR, we can observe that KSSHIBA tends to show a lower percentage of RVs to describe the kernel. This is due to the fact that KSSHIBA learns the relevance of each element and eliminates them accordingly, whereas KPCA and KCCA obtain this compact solutions with a random selection of RVs. 

\begin{table*}[h!b]
\centering
\caption{Results on the multitask databases for the automatic SV selection. The first subcolumn shows on the white subrow the mean and standard deviation of the R2 score and on the gray subrow the number of effective latent factors ($K_c$), the second subcolumn includes the percentage of RVs selected($\%RVs$).}
\label{tab:SupportVectors}
\begin{adjustbox}{max width=\linewidth}
\begin{tabular}{lccccccc}
\toprule
& \multicolumn{2}{c}{Sparse KSSHIBA} & \multicolumn{2}{c}{KPCA + LR} & \multicolumn{2}{c}{KCCA + LR} \\
& R2 - $K_c$ & $\%RVs$ & R2 - $K_c$ & $\%RVs$ & R2 - $K_c$ & $\%RVs$ \\
\midrule

\multirow{2}{*}{\textit{at1pd}} & $0.77 \pm 0.09$ & \multirow{2}{*}{$18.4 \pm 24.1$} & $0.78 \pm 0.09$ & \multirow{2}{*}{$69.7 \pm 32.9$} & $\mathbf{0.80 \pm 0.09}$ & \multirow{2}{*}{$84.8 \pm 27.5 $}\\
& \cellcolor{gray!10} $41 \pm 11$  & & \cellcolor{gray!10} $87 \pm 35$ & & \cellcolor{gray!10} $6$\\\midrule

\multirow{2}{*}{\textit{at7pd}} & $0.55 \pm 0.15$ & \multirow{2}{*}{$18.5 \pm 26.3$} & $0.56 \pm 0.18$ & \multirow{2}{*}{$79.7 \pm 31.7$} & $\mathbf{0.60 \pm 0.12}$ & \multirow{2}{*}{$73.9 \pm 34.1$}\\
& \cellcolor{gray!10} $70 \pm 27$ & & \cellcolor{gray!10} $90 \pm 37$  & & \cellcolor{gray!10} $6$\\ \midrule

 \multirow{2}{*}{\textit{oes97}} & $\mathbf{0.58 \pm 0.15}$ & \multirow{2}{*}{$38.6 \pm 24.5$} & $0.52 \pm 0.24$ & \multirow{2}{*}{$81.7 \pm 27.8$} & $0.42 \pm 0.30$ & \multirow{2}{*}{$23.9 \pm 27.8$}\\
& \cellcolor{gray!10} $61 \pm 7$ &  & \cellcolor{gray!10} $124 \pm 34$  &  & \cellcolor{gray!10} $16$ \\   \midrule

 \multirow{2}{*}{\textit{oes10}} & $\mathbf{0.77 \pm 0.11}$ & \multirow{2}{*}{$44.4 \pm 38.4$} & $0.71 \pm 0.12$ & \multirow{2}{*}{$71.9 \pm 11.6$} & $0.66 \pm 0.10$ & \multirow{2}{*}{$57.8 \pm 35.2$}\\
& \cellcolor{gray!10} $74 \pm 6$ &   & \cellcolor{gray!10} $132 \pm 53$ &  & \cellcolor{gray!10} $16$ \\ \midrule

 \multirow{2}{*}{\textit{edm}} & $\mathbf{0.42 \pm 0.21}$ & \multirow{2}{*}{$53.8 \pm 28.5$} & $0.41 \pm 0.26$ & \multirow{2}{*}{$52.5 \pm 30.5$} &  $0.20 \pm 0.14$ & \multirow{2}{*}{$22.7 \pm 13.6$}\\
& \cellcolor{gray!10} $13 \pm 4$ &  & \cellcolor{gray!10} $29 \pm 14$ &  & \cellcolor{gray!10} $2$ \\ \midrule

 \multirow{2}{*}{\textit{jura}} & $\mathbf{0.58 \pm 0.14}$ & \multirow{2}{*}{$48.7 \pm 38.4$} & $0.57 \pm 0.10$ & \multirow{2}{*}{$60.7 \pm 28.9$} & $0.36 \pm 0.09$ & \multirow{2}{*}{$18.9 \pm 7.5 $} \\
& \cellcolor{gray!10} $30 \pm 4$  &  & \cellcolor{gray!10} $59 \pm 14$ &  & \cellcolor{gray!10} $3$\\ \midrule

 \multirow{2}{*}{\textit{wq}} & $\mathbf{0.12 \pm 0.01}$ & \multirow{2}{*}{$58.1 \pm 33.2$} & $0.12 \pm 0.02$ & \multirow{2}{*}{$22.9 \pm 15.9$} & $0.10 \pm 0.01$ & \multirow{2}{*}{$5.9 \pm 3.1$}\\
& \cellcolor{gray!10} $21 \pm 2$ &  & \cellcolor{gray!10} $96 \pm 49$ &  & \cellcolor{gray!10} $14$ \\ \midrule
  
 \multirow{2}{*}{\textit{enb}} & $\mathbf{0.99 \pm 0.01}$ & \multirow{2}{*}{$19.5 \pm 12.8$} & $0.91 \pm 0.01$ & \multirow{2}{*}{$48.9 \pm 32.9$} & $0.97 \pm 0.01$ & \multirow{2}{*}{$41.9 \pm 12.2 $}\\
& \cellcolor{gray!10} $78 \pm 8$  &  & \cellcolor{gray!10} $28 \pm 1$ &  & \cellcolor{gray!10} $2$\\
\bottomrule
\end{tabular}
\end{adjustbox}
\end{table*}

\begin{figure}[h!t]
  \centering
  \begin{subfigure}[t]{0.48\linewidth}
    \centering
    \includegraphics[width=\linewidth]{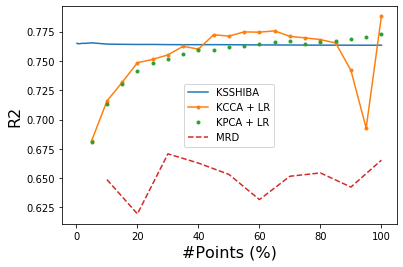}
    \caption{\textit{at1pd} database}
    \label{fig:sparse_atp1d}
  \end{subfigure}
  ~ 
  \begin{subfigure}[t]{0.48\linewidth}
    \centering
    \includegraphics[width=\linewidth]{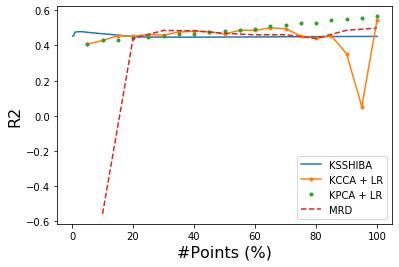}
    \caption{\textit{at7pd} database}
    \label{fig:sparse_oes97}
   \end{subfigure}
  \caption{R2 results with different percentages of RVs in KSSHIBA, KCCA+LR and KPCA+LR or inducing points in MRD.}
  \label{fig:sparse}
\end{figure}

To complete this analysis, Figure \ref{fig:sparse} includes the mean R2 over 10 folds varying the percentage of RVs. In this case, we show the results with the two databases for which SSHIBA is outperformed, the results with the other databases are available in \ref{sec:AppendixResults}. For the sake of comparison, we also included the MRD results when its percentage of inducing points is varied. 
On the one hand, we can observe that the MRD behavior w.r.t. to the number of inducing points is unstable and quite dependent on the database. Further, the position of inducing points uses a regular grid, since this implementation does not allow their optimization, causing high fluctuations in its performance.
 On the other hand, while KPCA+LR and KCCA+LR present fluctuations in their performance requiring to adjust the number of RVs to obtain an accurate performance, KSSHIBA has a relatively constant R2 value.  This phenomenon occurs because KSSHIBA learns the relevance of each SV and weight their influence on the update of the parameters during all the model inference. 


\subsection{Analysis of the feature relevance}

In order to test the feature relevance functionality (see Section \ref{sec:fl}),  we now study KSSHIBA on classification images database, using the image as input view and the category label as the output view, as this scheme provides an illustrative mechanism to analyze the feature (pixel) relevances. In particular, we used an aligned version of the Labeled Faces in the Wild (LFW) dataset \cite{LFWTech} obtained by \cite{wolf2010effective} and the  \textit{warpAR10P} ($60 \times 40$ pixels), \textit{Yale} ($32 \times 32$ pixels) and \textit{Olivetti} ($32 \times 32$ pixels) databases, which can be found in the Feature Selection Repository\footnote{\url{http://featureselection.asu.edu/datasets.php}}. Whereas the later databases were preprocessed, in LFW we had to crop the images to eliminate undesirable information and resize them to $60 \times 40$ pixels to reduce the computational cost of training the models; besides, to limit the size of the database we only used the images of the 7 people with most images in the database. The characteristics of these databases are  described in Table \ref{tab:Img-databases}.

\begin{table}[hb]
\renewcommand{\arraystretch}{1.2}
\caption{Characteristic of the faces databases used in this work.}
\label{tab:Img-databases}
\vspace{0.2cm}
\centering
\begin{adjustbox}{max width=\textwidth}
\begin{tabular}{cccc}
\toprule
Database & Samples & Features & Classes \\ \midrule
\textit{LFW}          & 1,277      & 2,400       & 7          \\ 
\textit{warpAR10P}    & 130        & 2,400       & 10          \\ 
\textit{Yale}         & 165        & 1,024       & 15         \\ 
\textit{Olivetti}     & 400        & 1,024       & 40         \\ 
\bottomrule
\end{tabular}
\end{adjustbox}
\end{table}

\begin{figure}[h!t]
  \centering
  \begin{subfigure}[t]{.3\linewidth}
    \centering
    \includegraphics[width=\linewidth]{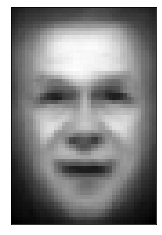}
    \caption{\textit{LFW}}
    \label{fig:mask_LFWA}
  \end{subfigure}
  ~
  \begin{subfigure}[t]{.381\linewidth}
    \centering
    \includegraphics[width=\linewidth]{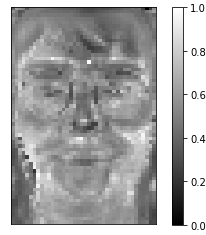}
    \caption{\textit{warpAR10P}}
    \label{fig:mask_warpAR10P}
  \end{subfigure}
  \\
    \begin{subfigure}[t]{.3\linewidth}
    \centering
    \includegraphics[width=\linewidth]{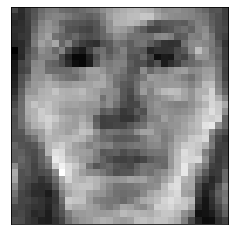}
    \caption{\textit{Yale}}
    \label{fig:mask_Yale}
  \end{subfigure}
    ~
    \begin{subfigure}[t]{.36\linewidth}
    \centering
    \includegraphics[width=\linewidth]{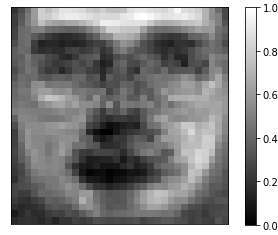}
    \caption{\textit{Olivetti}}
    \label{fig:mask_Olivetti}
  \end{subfigure}
  \caption{Feature masks learnt by the feature selection extension of KSSHIBA for different face recognition problems. The mask represent the importance of each pixel: lighter colours imply the pixel is more relevant while darker ones represent the pixel is less relevant.}
  \label{fig:Masks}
\end{figure}

The KSSHIBA scheme followed in this extension consist in including an ARD kernel in the input images to obtain a feature relevance analysis. This way,  Figure \ref{fig:Masks} shows the relevance mask  learnt by the model, over echa dataset, having a lighter colour for relevant pixels a darker colour when the pixel is not relevant. With these databases we can easily recognise the learnt face shape, which determine which pixels the model have to center on, paying less attention to the background and, in some cases, clearly defining the nose, cheek or chin, i.e., defining a mask of the face.

Analysing these masks we can see that the relevances learnt for \textit{LFW} is more informative and have a higher definition than the other, mainly because there is a considerably higher number of samples for this database. Both the \textit{warpAR10P} and the \textit{Olivetti} databases agree to focus on the area related to the glasses as there is representative number of images with glasses in both databases. Besides, we have that if one subject wears glasses in one image, he will also wear glasses in the rest, therefore having a relevant area to classify the different subjects in these datasets. Another fact that stands out for us is that the mouths and eyes are generally considered as non-relevant for the classification task, while mainly focusing on the hair, cheeks and the face shape. This is not so accentuated in the \textit{warpAR10P} dataset because there are a considerable number of images with cloths covering the face under the nose.



\subsection{Analysis of the extracted latent factors}

%
%

In this section we want to evaluate the interpretability of the extracted latent factors obtained by the proposed model in comparison to the MRD approach based on shared GPLVMs. In particular, we will analyse both of them on the \textit{Oil} classification database \citep{bishop1993analysis} (which has 2,000 samples, 12 features and 3 output classes). For this purpose we have trained both models with $15$ latent factors (number of features plus output classes) combined with ARD latent factor selection. KSSHIBA uses a RBF kernel only for the input view meanwhile MRD uses it for both their input and output views. Under these conditions, the accuracy in the prediction of the labels for the MRD was of $99.0\%$ and KSSHIBA achieved a $99.4 \%$. Furthermore, with the available MRD implementation (\textit{Matlab}), the computational time is not scalable for the number of data. 


\begin{figure}[h!t]
  \centering
  \begin{subfigure}[t]{0.4\linewidth}
    \centering
    \includegraphics[width=\linewidth]{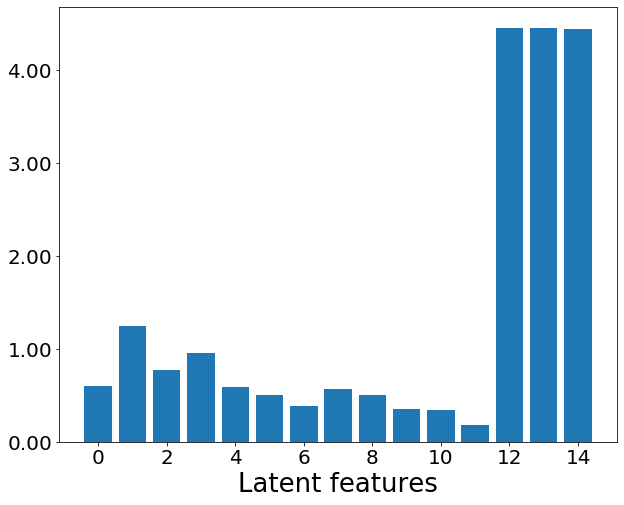}    
    \caption{MRD - common}
    \label{fig:GPLVM_Latents}
  \end{subfigure}
  \\
  \begin{subfigure}[t]{0.4\linewidth}
    \centering
  \includegraphics[width=\linewidth]{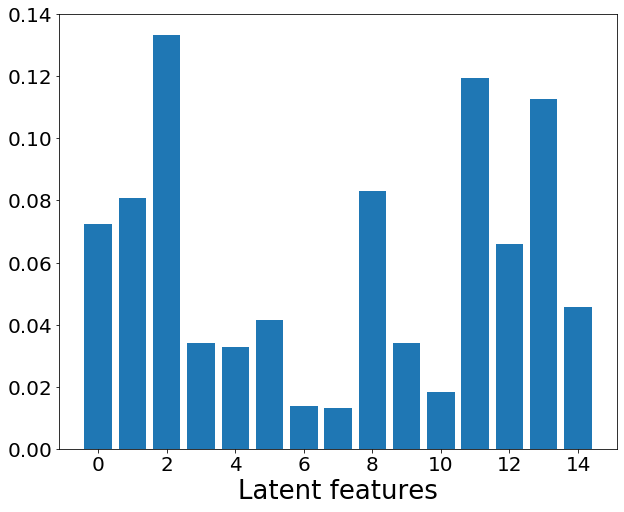}    
    \caption{KSSHIBA - input}
   \label{fig:KSSHIBA_input}
   \end{subfigure}
     ~ 
  \begin{subfigure}[t]{0.4\linewidth}
    \centering
  \includegraphics[width=\linewidth]{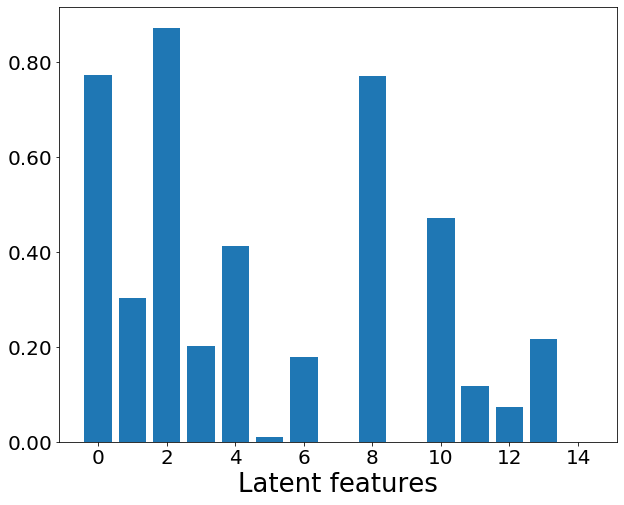}    
    \caption{KSSHIBA - output}
   \label{fig:KSSHIBA_output}
   \end{subfigure}
  \caption{Measure of relevance for each learnt latent factor on the \textit{Oil} database. Figure \ref{fig:GPLVM_Latents} shows the relevance of the commons for MRD model (all latents have resulted to be shared by both views). Figures \ref{fig:KSSHIBA_input} and \ref{fig:KSSHIBA_output} show, respectively, the relevance for the input view and the output view for KSSHIBA.}
 \label{fig:Oil_Latents}
\end{figure}
Figure \ref{fig:Oil_Latents} shows the learnt relevance for each latent factor for MRD ($W$) and KSSHIBA ($\gamm$). In the case of MRD, the formulation limits the model to specify the number of latent factors that are going to be used for the input and for the output data (Figure \ref{fig:GPLVM_Latents} shows the relevance for all these common factors). This means that the first 12 latent factors are related to the input view and the last 3 to the output, seeing that the model mainly focuses on the latter.
On the other hand, KSSHIBA presents independent weights for each view (see Figures \ref{fig:KSSHIBA_input} and \ref{fig:KSSHIBA_output}). This implies that the latent factors might be not relevant and could be pruned (latent 7), are only relevant for one view (latents 5 and 14) or are simultaneously relevant for more than one view (highlighting latents 0, 2 and 8). This latent information obtained by the inclusion of an ARD prior over the projection matrices $\Wm$, provides a more interpretable model.

\begin{figure}[h!t]
  \centering
  \begin{subfigure}[t]{0.4\linewidth}
    \centering
    \includegraphics[width=\linewidth]{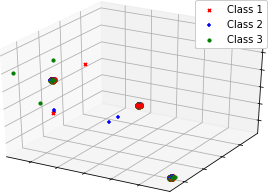}    
    \caption{Latent space of MRD}
    \label{fig:Oil_GP}
  \end{subfigure}
  ~ 
  \begin{subfigure}[t]{0.4\linewidth}
    \centering
  \includegraphics[width=\linewidth]{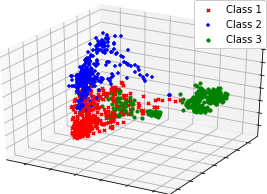}    
    \caption{Latent space of KSSHIBA}
  \label{fig:Oil_KSSHIBA}
  \end{subfigure}
  \caption{Learnt projections for the Oil database. Each figure shows the projections over the three most relevant factors: latents 12, 13 and 14 for MRD and latents 0, 2 and 8 for KSSHIBA.}
  \label{fig:Oil_Latents3D}
\end{figure}

Finally, to analyse the interpretability of the results, Figure \ref{fig:Oil_Latents3D} shows the most relevant extracted latent features with both models. For this three classes classification problem, we can see that the KSSHIBA algorithm is capable of finding a subspace where, with the three most relevant common latent factors, the classification problem is easily solved. Meanwhile, MRD with three latent factors projects most samples into a single point, needing more latent factors to discriminate the different classes. 

We think that MRD needs for a large number of common latent to obtain a discriminative space due to the inclusion of a non-linearity in its output view. In fact, this suggests that using a simple linear output view in KSSHIBA facilities the achievement of not only a better performance, but also a more informative and discriminative latent space.

\subsection{Multiple kernel learning for factor analysis}

One of the main functionalities of KSSHIBA is its capacity to combine multiple views into a single model. We can take advantage of this property when reconstructing kernel representations to combine different types of kernels (one per view) and  obtain a latent representation resulting from the linear combination of different kernels, known as MKL.
Furthermore, the nature of the algorithm allows us to automatically learn the relevance of each kernel while combining different kernels in one model.

To analyse this functionality, we decided to test it with three different databases: \textit{Arrhythmia} and \textit{Landsat} from the UCI repository \citep{UCI:2019} and \textit{Fashion MNIST} \citep{FMNIST}. These databases are described in Table \ref{tab:MKL_database}.
\begin{table}[hb]
\renewcommand{\arraystretch}{1.2}
\caption{Characteristic of the databases used for MKL.}
\label{tab:MKL_database}
\vspace{0.2cm}
\centering
\begin{adjustbox}{max width=\textwidth}
\begin{tabular}{ccccc}
\toprule
Database & Samples & \multicolumn{2}{c}{Features} & Classes \\
 & & Kernel 1 & Kernel 2 \\
\midrule
\textit{Arrhythmia}          & 452      & 15  & 264   & 2          \\ 
\textit{Landsat}             & 6,435    & 4   & 4     & 6        \\ 
\textit{Fashion MNIST}       & 1,000    & 784 & 784   & 10         \\ 
\bottomrule
\end{tabular}
\end{adjustbox}
\end{table}

To analyze the MKL KSSHIBA performance, we have compare it with a SVM classifier and a CCA to extract the latent factors and then a SVM to classify (CCA+SVM), using as input a linear combination of two different kernels:
\begin{align}
    K\p*{X_{i,:}, X_{j,:}} = \mu K^1\p*{X_{i,S}, X_{j,S}} + (1-\mu) K^2\p*{X_{i,R}, X_{j,R}} \label{eq:Kernel}
\end{align}
where $K^1$ is the first kernel, $K^2$ is the second kernel and $\mu \in [0,1]$ is the combination coefficient. 
Using these equations for the \textit{Arrhythmia} database, we defined $K^1$ as a RBF kernel with the demographics and general ECG information (15 features) and $K^2$ as a RBF kernel with the channel parameters like width and amplitude (263 features). For the \textit{Landsat} database we defined $K^1$ as a RBF kernel with the spectral information (4 features) while $K^2$ was a RBF kernel uses the contextual information (4 features), as proposed in \citep{amoros2011multitemporal}. Finally, for the \textit{Fashion MNIST} database we used only 1,000 samples for the experiment, with 784 features, using the same data for each kernel, setting $K^1$ to be RBF and $K^2$ to be polynomial.
We validated the $C$ and $\gamma$ parameters for the SVM as well as the combination coefficient $\mu$. For CCA we set the number of latent factors to $\#classes-1$.

\begin{table}[hb]
\renewcommand{\arraystretch}{1.2}
\caption{Multiple Kernel Learning analysis. The performance measure used is the multiclass AUC, which we calculated in a 10-fold cross-validation.}
\label{tab:MKL}
\vspace{0.2cm}
\centering
\begin{adjustbox}{max width=\textwidth}
\begin{tabular}{ccccccccc}
\toprule
Database & MKL SSHIBA & MKL SSHIBA + RVS & SVM & CCA+SVM \\\midrule
\textit{Arrhythmia} & $0.809 \pm 0.057$ & $\bm{0.818 \pm 0.046}$ & $0.774 \pm 0.045$ & $0.737 \pm 0.059$         \\ 
\textit{Landsat}    &  $\bm{0.983 \pm 0.002}$ & $\bm{0.983 \pm 0.002}$ & $0.963 \pm 0.042$ & $0.980 \pm 0.002$         \\ 
\textit{Fashion MNIST}    &  $0.951 \pm 0.011$ & $0.952 \pm 0.013$ & $\bm{0.970 \pm 0.010}$ & $0.844 \pm 0.012$         \\ 
\bottomrule
\end{tabular}
\end{adjustbox}
\end{table}

The performance obtained with these configurations are included in Table \ref{tab:MKL}, where we used two versions of KSSHIBA, with and without RVs Selection (RVS), and multiclass AUC as the performance measure. The results prove that the combination of the different kernels carried out by MKL-SSHIBA is capable of providing good results with respect to the baselines. In particular, in the \textit{Arrhythmia} dataset we obtained an improvement of up to $0.044$ with respect to the SVM. Furthermore, we can see that the inclusion of the RV selection provides a more robust model, as it automatically selects the RVs that are relevant for the problem. Although in the \textit{Fashion MNIST} database the SVM provides a better result than the proposed model, we consider that the improvement is not enough to justify the lack of feature extraction; in fact, the results obtained by SVM with feature extraction are considerably lower than the obtained by the proposed model.

To further analyse the capabilities of the presented model, we decided to use the \textit{Fashion MNIST} database with three different kernels to analyse the learnt latent factors. In particular, we decided to combine the RBF and polynomial with a linear kernel. We included the labels in another view and used them as an output view. The performance obtained with this framework is equivalent to the one previously explored, probing that the combination of different kernels do not deteriorate the performance of the model.

\begin{figure}[ht]
 \centering
 \includegraphics[page=1,width=\linewidth]{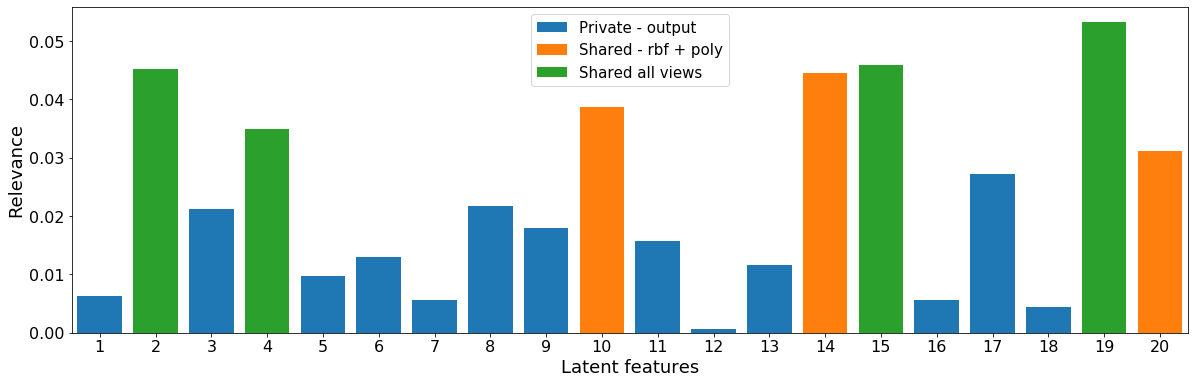}\\
 \captionof{figure}{Relevance of the learnt latent factors. Measure of relevance on \textit{Fashion MNIST} database combining kernels and labels on different views.}
  \label{fig:multikernel}
\end{figure}

Figure \ref{fig:multikernel} shows the relevance of each latent factor in the joint scenario (all kernels used).  From the original $100$ latent factors, the RBF and the polynomial kernels share $3$ and the output view only uses $17$, being most of them private. In fact, we can observe that the improved performance of the model is obtained using only four common factors to all views for the prediction of the output label. Note that, only the private output and the shared factors are used for the prediction, while the rest capture relations between views.



\section{Conclusions}
\label{sec:conclusion}

We propose a novel probabilistic latent variable model to generate kernel relationships, instead of data observations, based on a linear generative model. We introduce this model using the Bayesian Inter-Battery Factor Analysis (BIBFA) approach proposed in \citep{sevilla2020sparse} to show its capabilities to efficiently face semi-supervised heterogeneous multi-view problems combining linear and non-linear data representations. Besides, we extend the model formulation to provide the automatic selection of RVs, obtaining scalable solutions, as well as include an ARD prior over the kernel to obtain the feature relevance functionality. The model performance is evaluated in multi-dimensional regression, feature relevance over images and multiple-kernel learning problems demonstrating that the inclusion of kernelized observations provide fruitful results.

The results prove the relevance of the proposed formulation, achieving a not only competitive performance, but also transforming the data to a reduced set of interpretable latent variables and a compact model consisting in a reduced subset of RVs. Furthermore, the feature relevance criteria is able to learn relevant masks which provide insight knowledge of the input space for the goal task. Finally, we also proved that taking advantage of the multiview nature of KSSHIBA we can easily combine different kernels to enhance, even more, the final performance of the model. This model contributes to the field of Artificial Intelligence by providing a model capable of adapting to a wide range of scenarios. In particular, the ability of combining feature selection, with heterogeneous data, imputation of missing values and combination of multiple kernels can be very useful for some biomedic or face recognition problems.  In these, there usually are a considerable number of missing values, the interpretability given by the feature relevance and the learnt latent factors might improve the analysis of the results and MKL allows the model to combine linear and non-linear data.



\section*{Acknowledgments}
\label{sec:acknowledgments}
The work of Pablo M. Olmos is supported by   Spanish  government   MINECO   under   grant  RTI2018-099655-B-10,  by  Comunidad de   Madrid   under   grants   IND2017/TIC-7618, IND2018/TIC-9649, and Y2018/TCS-4705,  by BBVA Foundation under the Deep-DARWiN project,  and   by   the European  Union  (FEDER and the European Research Council (ERC) through the European Unions Horizon 2020 research and innovation program under Grant 714161). C. Sevilla-Salcedo and V. G{\'o}mez-Verdejo's work has been partly funded by the Spanish MINECO grant TEC2017-83838-R. The work of Alejandro Guerrero is supported by MINECO and EU FEDER.

\bibliography{bibliography}

\appendix
\section{SSHIBA's summary}
\label{sec:AppendixSSHIBA}

Semi-supervised Sparse Heterogeneous Inter-battery Bayesian Analysis (SSHIBA) is based on Bayesian Inter-Battery Factor Analysis (BIBFA) presented in \citep{klami2013bayesian}. The main goal of both is to jointly project different data representations, defined as ``views'', into a discriminative low-dimensional space. The joint probability density function of BIBFA can be defined as
\begin{align}
    \Zn \eqsimil \N(0,I_{K_c}) \label{eq:Zprior}\\
	\Wkm \eqsimil \N\p*{0,\p*{\akm}^{-1} I_{K_c}} \label{eq:Wprior}\\
    \Xnm | \Zn \eqsimil \N(\Zn \WmT, \taum^{-1} I_{D_m}) \label{eq:Xprior}\\
    \alpha_{k}^{(m)} \eqsimil \Gamma\p*{a^{\alpha^{(m)}}, b^{\alpha^{(m)}}} \label{eq:Alphaprior}\\
	\taum \eqsimil \Gamma\p*{ a^{\taum}, b^{\taum} } \label{eq:Tauprior}
\end{align}

\begin{figure}[h!]
  \centering
  \hspace{-1.5cm}
  \begin{subfigure}[t]{0.4\textwidth}
    \centering
    \includegraphics[width=0.9\textwidth]{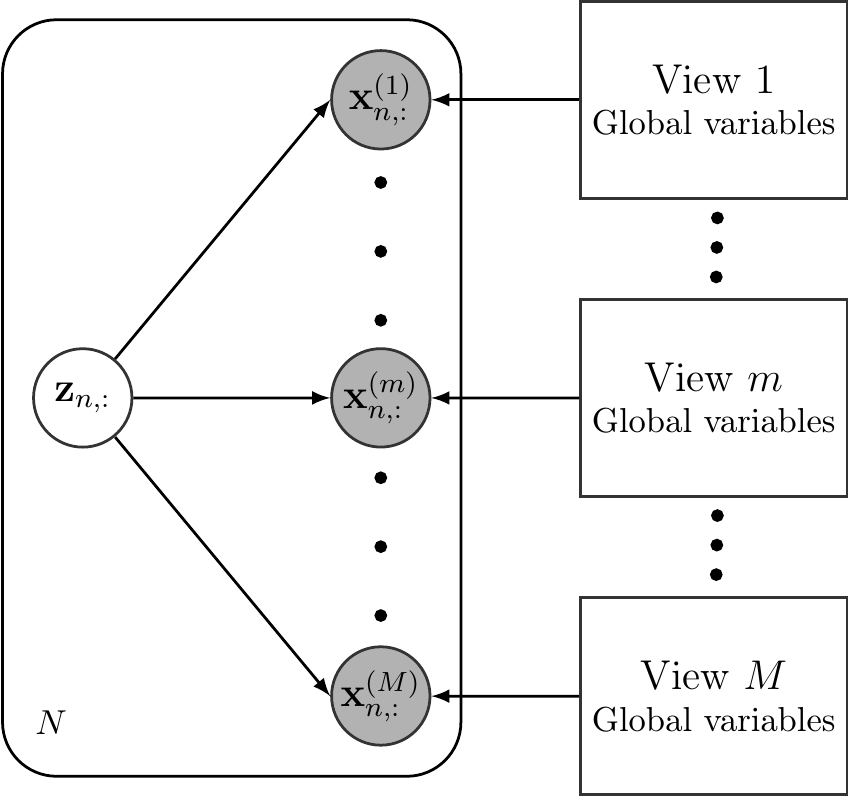}\\
    \caption{Multi-view model.}
    \label{fig:BIBFASchemea}
  \end{subfigure}
  ~
  \begin{subfigure}[t]{0.4\textwidth}
    \centering
    \includegraphics[width=1\linewidth]{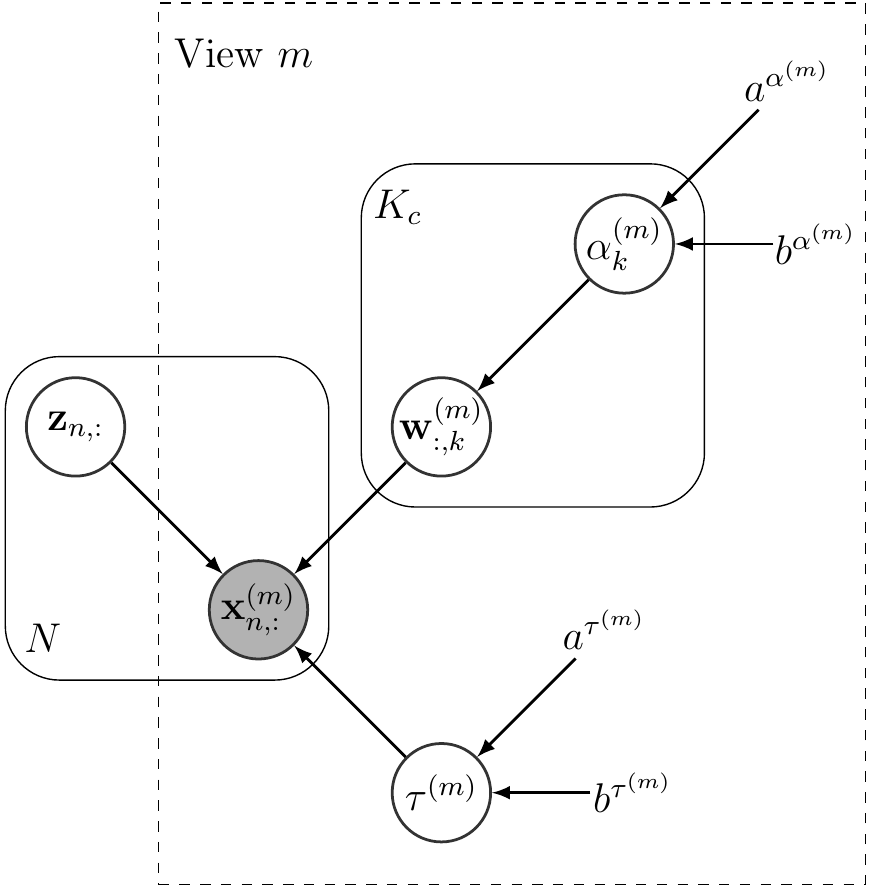}
    \caption{Zoom in view m.}
    \label{fig:BIBFASchemeb}
  \end{subfigure}
  \caption{Plate diagram for the BIBFA and SSHIBA for real-data graphical model. Gray circles denote observed variables, white circles unobserved random variables. The nodes without a circle correspond to the hyperparameters.}
  \label{fig:BIBFAScheme}
\end{figure}

The graphic model associated to both is included in Figure \ref{fig:BIBFAScheme}. However, SSHIBA presents certain updates to this formulation which allow the model to adapt to more scenarios \cite{sevilla2020sparse}. In particular, it is able to combine semi-supervised learning with feature selection while being able to model categorical and multi-dimensional binary (multi-label) data, besides real data.

\section{Extended experiments}
\label{sec:AppendixResults}
In this section we present a more extensive version of the results presented in the article. These results include some databases and baselines that we did not include due to the space limitations or the lack of relevance of the results. In particular, Table \ref{tab:MultiTask_sup} includes KCCA and a MLP where the second hidden layer works as a feature extractor, including a bottleneck of $C$.


\begin{table}[h!p]
\renewcommand{\arraystretch}{1.2}
\caption{Results on the multitask databases of the KSSHIBA and the different methods under study. In this case the data is normalised and the kernel, if applied, is centred. Each subrow represents the mean and standard deviation of the R2 score (white) and latent factor (light gray) used, respectively.}
\label{tab:MultiTask_sup}
\vspace{0.2cm}
\centering
\begin{adjustbox}{max width=\textwidth}
\begin{tabular}{cccccccccc}
\toprule
 & \multirow{2}{*}{KSSHIBA}  & KSSHIBA & \multirow{2}{*}{MRD} & \multirow{2}{*}{KPCA + LR} & \multirow{2}{*}{KCCA} & \multirow{2}{*}{KCCA + LR} & \multirow{2}{*}{SVR rbf} & MLP & \multirow{2}{*}{MLP}  \\
 &   & $K_c = C$ &  &  &  &  &  & $K_c = C$  &   \\ \midrule
\multirow{2}{*}{\textit{at1pd}} & $0.77 \pm 0.09$ & $\mathbf{0.78 \pm 0.09}$ & $0.67 \pm 0.07$ & $0.67 \pm 0.12$ & $0.45 \pm 0.05$ & $0.75 \pm 0.11$ & \multirow{2}{*}{$0.01 \pm 0.05$} & $0.75 \pm 0.09$  & \multirow{2}{*}{$0.77 \pm 0.12$} \\
& \cellcolor{gray!10} $53 \pm 8$ & \cellcolor{gray!10} $6$ & \cellcolor{gray!10} $12$ & \cellcolor{gray!10} $22 \pm 10$ & \cellcolor{gray!10} $6$ & \cellcolor{gray!10} $6$ & & \cellcolor{gray!10} $6$\\ \midrule
 
\multirow{2}{*}{\textit{at7pd}} & $0.48 \pm 0.26$ & $0.52 \pm 0.13$ & $0.48 \pm 0.12$ & $0.39 \pm 0.19$ & $0.24 \pm 0.05$ & $\mathbf{0.57 \pm 0.16}$ & \multirow{2}{*}{$0.01 \pm 0.03$} & $0.29 \pm 0.33$ & \multirow{2}{*}{$0.35 \pm 0.69$} \\
& \cellcolor{gray!10} $53 \pm 11$ & \cellcolor{gray!10} $6$ & \cellcolor{gray!10} $12$ & \cellcolor{gray!10} $21 \pm 1$ & \cellcolor{gray!10} $6$ & \cellcolor{gray!10} $6$ & & \cellcolor{gray!10} $6$\\ \midrule

 \multirow{2}{*}{\textit{oes97}} & $0.63 \pm 0.16$ & $\mathbf{0.69 \pm 0.10}$ & $0.34 \pm 0.07$ & $0.45 \pm 0.20$ & $0.30 \pm 0.08$ & $0.36 \pm 0.09$ & \multirow{2}{*}{$0.39 \pm 0.10$} & $0.57 \pm 0.22$ & \multirow{2}{*}{$0.58 \pm 0.21$} \\
& \cellcolor{gray!10} $108 \pm 11$ & \cellcolor{gray!10} $16$ & \cellcolor{gray!10} $32$ & \cellcolor{gray!10} $12 \pm 7 $ & \cellcolor{gray!10} $16$ & \cellcolor{gray!10} $16$ & & \cellcolor{gray!10} $ 16$\\ \midrule

 \multirow{2}{*}{\textit{oes10}} & $0.79 \pm 0.08$ & $\mathbf{0.80 \pm 0.07}$ & $0.38 \pm 0.07$ & $0.59 \pm 0.15$ & $0.35 \pm 0.17$ & $0.43 \pm 0.12$ & \multirow{2}{*}{$0.47 \pm 0.12$} & $0.77 \pm 0.07$ & \multirow{2}{*}{$0.76 \pm 0.08$} \\
 & \cellcolor{gray!10} $104 \pm 22$ & \cellcolor{gray!10} $16$ & \cellcolor{gray!10} $32$ & \cellcolor{gray!10} $14 \pm 7$ & \cellcolor{gray!10} $16$ & \cellcolor{gray!10} $16$ & & \cellcolor{gray!10} $16$\\ \midrule

 \multirow{2}{*}{\textit{edm}} & $0.37 \pm 0.19$ & $0.21 \pm 0.09$ & $-0.17 \pm 0.45$ & $\mathbf{0.38 \pm 0.19}$ & $0.26 \pm 0.18$ & $0.18 \pm 0.26$ & \multirow{2}{*}{$0.35 \pm 0.19$} & $0.14 \pm 0.17$ & \multirow{2}{*}{$0.26 \pm 0.21$}\\
 & \cellcolor{gray!10} $17 \pm 2$ & \cellcolor{gray!10} $2$ & \cellcolor{gray!10} $4$ & \cellcolor{gray!10} $16 \pm 5$ & \cellcolor{gray!10} $2$ & \cellcolor{gray!10} $2$ & & \cellcolor{gray!10} $2$\\ \midrule

 \multirow{2}{*}{\textit{jura}} & $\mathbf{0.61 \pm 0.10}$ & $0.30 \pm 0.10$ & $0.57 \pm 0.06$ & $0.38 \pm 0.11$ & $0.11 \pm 0.08$ & $0.18 \pm 0.15$ & \multirow{2}{*}{$0.60 \pm 0.05$} & $0.32 \pm 0.12$ & \multirow{2}{*}{$\mathbf{0.61 \pm 0.06}$}\\
 & \cellcolor{gray!10} $64 \pm 7$ & \cellcolor{gray!10} $3$ & \cellcolor{gray!10} $6$ & \cellcolor{gray!10} $23 \pm 1$ & \cellcolor{gray!10} $3$ & \cellcolor{gray!10} $3$ & & \cellcolor{gray!10} $3$\\ \midrule

 \multirow{2}{*}{\textit{wq}} & $0.12 \pm 0.01$ & $0.12 \pm 0.01$ & $-0.35 \pm 0.08$ & $0.09 \pm 0.02$ & $-0.01 \pm 0.01$ & $-0.01 \pm 0.01$ & \multirow{2}{*}{$0.08 \pm 0.02$} & $0.10 \pm 0.02$ & \multirow{2}{*}{$\mathbf{0.13 \pm 0.03}$}\\
 & \cellcolor{gray!10} $48 \pm 3$ & \cellcolor{gray!10} $14$ & \cellcolor{gray!10} $28$ & \cellcolor{gray!10} $29 \pm 0.98$ & \cellcolor{gray!10} $14$ & \cellcolor{gray!10} $14$ & & \cellcolor{gray!10} $14$\\ \midrule

 \multirow{2}{*}{\textit{enb}} & $\mathbf{0.99 \pm 0.01}$ & $0.86 \pm 0.02$ & $0.91 \pm 0.01$ & $0.86 \pm 0.01$ & $0.96 \pm 0.01$ & $0.98 \pm 0.01$ & \multirow{2}{*}{$\mathbf{0.99 \pm 0.01}$} & $0.89 \pm 0.01$ & \multirow{2}{*}{$\mathbf{0.99 \pm 0.08}$}\\
 & \cellcolor{gray!10} $118 \pm 4$ & \cellcolor{gray!10} $2$ & \cellcolor{gray!10} $4$ & \cellcolor{gray!10} $13 \pm 1$ & \cellcolor{gray!10} $2$ & \cellcolor{gray!10} $2$ & & \cellcolor{gray!10} $2$\\ \bottomrule
\end{tabular}
\end{adjustbox}
\end{table}

\begin{figure}[h!t]
  \centering
     \begin{subfigure}[t]{0.45\textwidth}
     \centering
     \includegraphics[width=\textwidth]{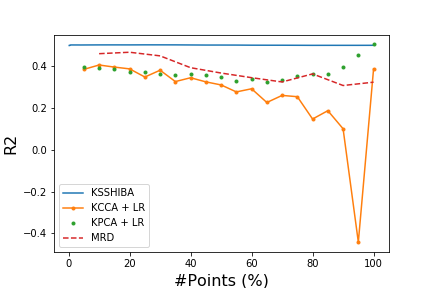}
     \caption{\textit{oes97} database.}
     \label{fig:sparse_oes97_sup}
   \end{subfigure}
   ~ 
   \begin{subfigure}[t]{0.45\textwidth}
     \centering
     \includegraphics[width=\textwidth]{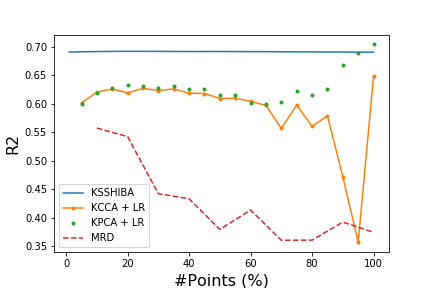}
     \caption{\textit{oes10} database.}
     \label{fig:sparse_oes10_sup}
   \end{subfigure}
   \\
     \begin{subfigure}[t]{0.45\textwidth}
     \centering
     \includegraphics[width=\textwidth]{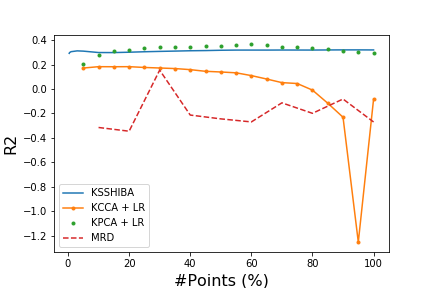}
     \caption{\textit{edm} database.}
     \label{fig:sparse_edm_sup}
   \end{subfigure}
   ~ 
   \begin{subfigure}[t]{0.45\textwidth}
     \centering
     \includegraphics[width=\textwidth]{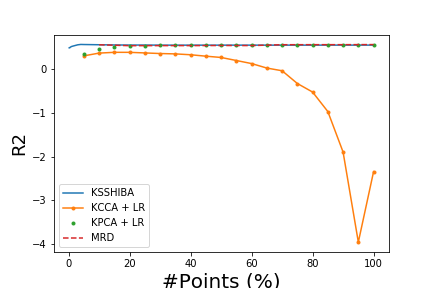}
     \caption{\textit{jura} database.}
     \label{fig:sparse_jura_sup}
   \end{subfigure}
   \\
     \begin{subfigure}[t]{0.45\textwidth}
     \centering
     \includegraphics[width=\textwidth]{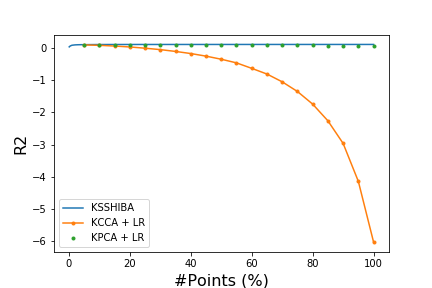}
     \caption{\textit{wq} database.}
     \label{fig:sparse_wq_sup}
   \end{subfigure}
   ~ 
   \begin{subfigure}[t]{0.45\textwidth}
     \centering
     \includegraphics[width=\textwidth]{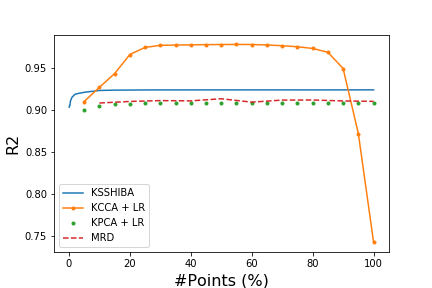}
     \caption{\textit{enb} database.}
     \label{fig:sparse_enb_sup}
   \end{subfigure}
  \caption{R2 results with different percentages of RVs in the the KSSHIBA, KCCA+LR and KPCA+LR or inducing points in MRD.}
  \label{fig:sparse_sup}
\end{figure}

Finally, Figure \ref{fig:sparse_sup} presents the analysis of the effect of the number of RVs or inducing points (MRD) analysed in the evaluation of the solution in terms of RVs. In particular, we include here the results obtained for the databases not included in the main article. The results on MRD are not included for the \textit{wq} database because the model iterations have not ended at the moment this material is done due to the high computational time required by the library.

\end{document}